\documentclass[letterpaper,journal]{IEEEtran}
\usepackage{amsmath,amsfonts}
\usepackage{algorithmic}
\usepackage{algorithm}
\usepackage{array}
\usepackage{booktabs}
\usepackage[caption=false,font=normalsize,labelfont=sf,textfont=sf]{subfig}
\usepackage{textcomp}
\usepackage{multirow}
\usepackage{stfloats}
\usepackage{colortbl}
\usepackage{xcolor}
\usepackage{url}
\usepackage{verbatim}
\usepackage{graphicx}
\usepackage{cite}
\begin{document}

\title{DecoKAN: Interpretable Decomposition for Forecasting Cryptocurrency Market Dynamics}

\author{
        Yuan~Gao,
        Zhenguo~Dong,
        Xuelong~Wang,
        Zhiqiang~Wang,
        Yong~Zhang,~\IEEEmembership{Member,~IEEE,}
        and Shaofan~Wang
\thanks{This work was supported by the Beijing Municipal Natural Science Foundation (No. L251067), the Fundamental Research Funds for the Central Universities (No. 3282024058, 3282024052), and the China Postdoctoral Science Foundation (No. 2019M650608). (Corresponding author: Zhiqiang Wang.)}
\thanks{Yuan Gao, Zhenguo Dong, Xuelong Wang, and Zhiqiang Wang are with the Beijing Electronic Science and Technology Institute, Beijing 100070, China. Yong Zhang and Shaofan Wang are with the Beijing University of Technology, Beijing 100124, China. (E-mail: gy@besti.edu.cn; dongzgxx@163.com; 20243806@mail.besti.edu.cn; wangzq@besti.edu.cn; zhangyong2010@bjut.edu.cn; wangshaofan@bjut.edu.cn).}
}



\maketitle

\begin{abstract}
Accurate and interpretable forecasting of multivariate time series is crucial for understanding the complex dynamics of cryptocurrency markets in digital asset systems.
Advanced deep learning methodologies, particularly Transformer-based and MLP-based architectures, have achieved competitive predictive performance in cryptocurrency forecasting tasks.
However, cryptocurrency data is inherently composed of long-term socio-economic trends and local high-frequency speculative oscillations. 
Existing deep learning-based 'black-box' models fail to effectively decouple these composite dynamics or provide the interpretability needed for trustworthy financial decision-making.
To overcome these limitations, we propose DecoKAN, an interpretable forecasting framework that integrates multi-level Discrete Wavelet Transform (DWT) for decoupling and hierarchical signal decomposition with Kolmogorov–Arnold Network (KAN) mixers for transparent and interpretable nonlinear modeling.
The DWT component decomposes complex cryptocurrency time series into distinct frequency components, enabling frequency-specific analysis, while KAN mixers provide intrinsically interpretable spline-based mappings within each decomposed subseries. 
Furthermore, interpretability is enhanced through a symbolic analysis pipeline involving sparsification, pruning, and symbolization, which produces concise analytical expressions offering symbolic representations of the learned patterns.
Extensive experiments demonstrate that DecoKAN achieves the lowest average Mean Squared Error on all tested real-world cryptocurrency datasets (BTC, ETH, XMR), consistently outperforming a comprehensive suite of competitive state-of-the-art baselines.
These results validate DecoKAN’s potential to bridge the gap between predictive accuracy and model transparency, advancing trustworthy decision support within complex cryptocurrency markets.
\end{abstract}


\section{Introduction}
\IEEEPARstart{C}{ryptocurrency} systems generate massive volumes of time series data by continuously recording observations and events over extended periods. Accurate forecasting of such data has become essential for optimizing investment strategies, managing market risks, and maintaining economic stability. For example, forecasting trading volume enables better liquidity management to cope with market panic, while predicting price volatility contributes to the design of more resilient decentralized finance (DeFi) protocols \cite{zhu2019controllable,shen2023mlsurvey}. The complex nature of cryptocurrency markets, effectively functioning as large-scale computational social systems driven by heterogeneous agent interactions \cite{zhang2024novel}, characterized by interdependencies among variables and dynamics across multiple temporal scales, necessitates robust and trustworthy forecasting methodologies.

Early efforts in time series forecasting primarily relied on traditional statistical methods. Models from the Autoregressive Integrated Moving Average (ARIMA) family, including seasonal variants such as SARIMA and extensions incorporating exogenous variables like ARIMAX, established a strong statistical foundation for time series forecasting \cite{box2015time}. These statistical approaches offered simplicity and, importantly, high interpretability through transparent mathematical formulations that explicitly model trend, seasonality, and linear dependencies. In parallel, conventional machine learning techniques such as Hidden Markov Models (HMMs) were also explored, providing a probabilistic framework to better capture dynamic non-linear behaviors in financial time series \cite{hassan2005stock}. However, both traditional statistical and early machine learning models often struggled to represent the highly non-linear patterns, abrupt shifts, long-range dependencies, and intricate cross-variable relationships typical of large-scale financial time series, particularly within the volatile and dynamic digital financial market \cite{katsiampa2017volatility}.

\begin{figure}[!t]
\centering
\includegraphics[width=0.5\textwidth]{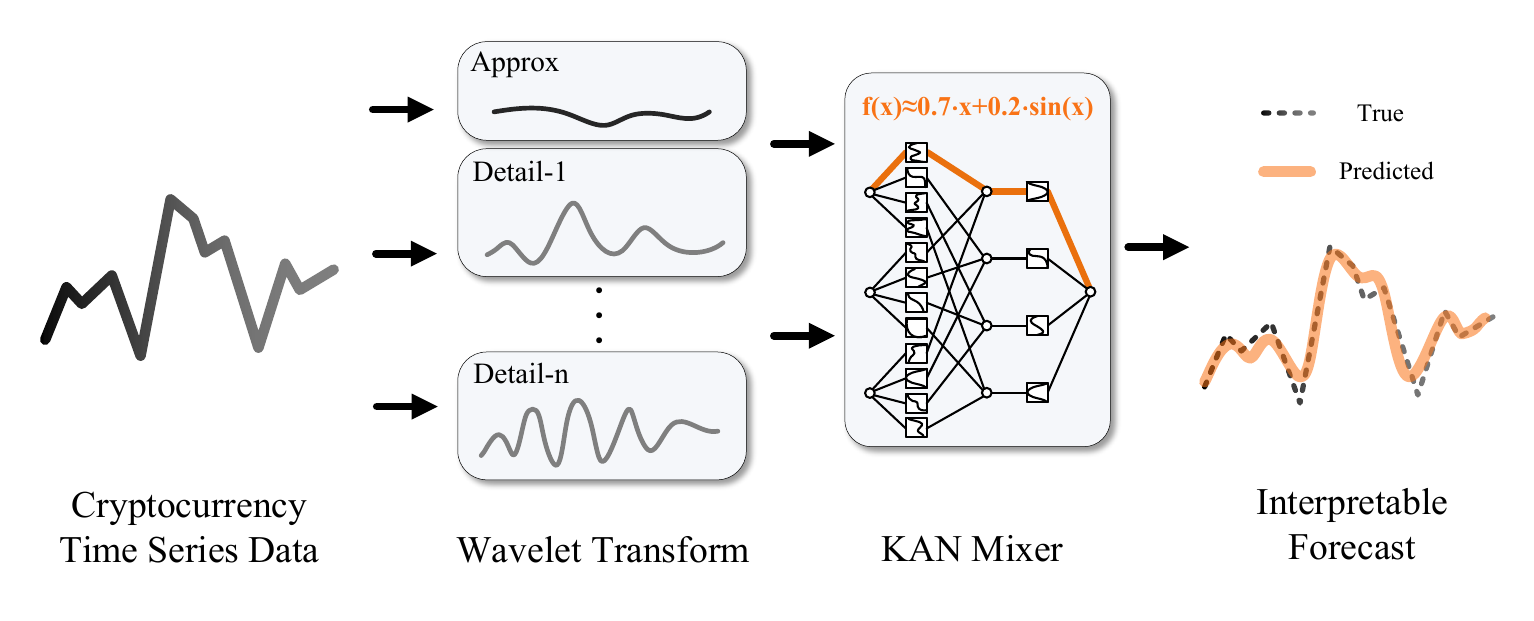}
\caption{Conceptual overview of the DecoKAN framework for interpretable time series forecasting.}
\label{fig_1}
\end{figure}

Deep learning methods have subsequently emerged as powerful tools for addressing these challenges, leveraging their strong nonlinear modeling capabilities. Early approaches employed Recurrent Neural Networks (RNNs), such as Long Short-Term Memory (LSTM) networks, and Convolutional Neural Networks (CNNs) adapted for sequential data \cite{liu2022scinet}.
Later, Transformer-based architectures \cite{vaswani2017attention} became dominant due to their ability to model long-term dependencies, with models like Informer \cite{zhou2021informer}, Autoformer \cite{wu2021autoformer}, and FEDformer \cite{zhou2022fedformer} achieving notable success, while Non-stationary Transformers \cite{liu2023nonstationary} aimed to mitigate distributional shifts. 
In the specific context of cryptocurrency trading, researchers have further tailored these architectures to address domain-specific challenges. For instance, recent works have enhanced Transformer models with market sentiment indices with advanced market sentiment analysis \cite{chiong2023novel} to improve prediction accuracy, and integrated diverse data sources—combining financial, on-chain, and social media metrics—to capture complex market dynamics \cite{gurgul2025deep}.
More recent architectures, including PatchTST\cite{nie2023time} and iTransformer\cite{liu2023itransformer}, further optimized Transformer efficiency and locality. In parallel, motivated by studies questioning the necessity of complex attention mechanisms for all forecasting tasks , several simpler yet competitive MLP-based architectures have been proposed, including DLinear\cite{zeng2023transformers}, TimeMixer\cite{wang2023timemixer}, TSMixer\cite{chen2023tsmixer} and WPMixer\cite{murad2025wpmixer}. These models often employ decomposition techniques, such as moving averages (TimeMixer) or wavelet transforms (WPMixer), to improve predictive robustness. Other CNN-based models such as TimesNet \cite{wu2023timesnet} and TimeXer \cite{TimeXer} extend the MLP-Mixer paradigm \cite{tolstikhin2021mlpmixer}, further enriching the design space. 
Collectively, these developments have substantially advanced the state-of-the-art in forecasting accuracy. However, directly applying these Transformer-based and MLP-based architectures to cryptocurrency price prediction remains problematic. These general-purpose models often struggle to adapt to the idiosyncrasies of cryptocurrency data, such as its extreme volatility, regime-switching non-stationarity, and the intricate mixture of long-term adoption trends with high-frequency speculative noise. Consequently, applying these sophisticated models in this context faces two fundamental challenges.

\textbf{Challenge 1:} Decoupling and Modeling Composite Signals. 
Cryptocurrency time series display complex temporal patterns that combine long-term structural trends with short-term volatility and noise. Explicitly decomposing and modeling these distinct components separately, rather than as a mixed signal, remains a challenge for most conventional forecasting architectures.

\textbf{Challenge 2:} Lack of interpretability. 
Although modern deep learning models achieve strong predictive accuracy, they often operate as opaque black boxes, offering little insight into their internal reasoning. This lack of transparency undermines trust and hinders adoption, particularly in the high-stakes financial contexts of DeFi and digital assets.

A review of existing approaches reveals both substantial progress and persistent limitations. Traditional statistical methods are interpretable but fail to capture the nonlinear complexity of market dynamics. Advanced deep learning models deliver higher accuracy, and certain architectures, such as WPMixer \cite{murad2025wpmixer}, employ decomposition strategies to decouple the composite signals. Nevertheless, while these decomposition-based frameworks separate the components, they still rely on opaque black boxes for modeling.
For signal decomposition, the Discrete Wavelet Transform (DWT) provides an established and effective method of separating signals into distinct components. At the same time, the persistent black-box nature of high-performance models continues to impede transparency and auditability, which are essential for trust and risk management in financial decision-making systems. Recently, the Kolmogorov-Arnold Network (KAN) paradigm has emerged as a promising direction, offering intrinsic interpretability through learnable spline activations and explicit symbolic representations \cite{liu2025kan, liu2024kan}.
The potential of applying the combination of KAN's symbolic transparency and DWT's proven capability in decomposing multivariate time series data to cryptocurrency market data remains unexplored. 

To bridge this gap, we propose DecoKAN (conceptually illustrated in Fig. 1), a novel interpretable forecasting framework that synergistically combines the hierarchical decomposition capabilities of DWT with the transparent, non-linear modeling power of KANs. Fig. 1 illustrates the core concept: cryptocurrency time series data is decomposed by wavelet transform, processed by a KAN-based mixer, and reconstructed into an interpretable forecast. This design enables accurate forecasting of complex market dynamics while maintaining transparent internal reasoning and scrutable logic.
Our contributions can be summarized as follows:

(1) In this paper, a novel framework named DecoKAN is proposed, which applies multi-level Discrete Wavelet Transform (DWT) to decompose composite financial time series into distinct approximation and detail components. Each component is then processed independently within dedicated Resolution Branches, thereby isolating the processing of distinct frequency bands and allowing for specialized pattern learning.

(2) A Kolmogorov–Arnold Network (KAN)-based mixer is introduced to address the opacity of conventional architectures. By replacing standard MLP layers with learnable spline activation functions, this mixer provides intrinsic interpretability and enables precise, transparent modeling of the relationships specific to each decoupled component, preserving the unique information captured by the wavelet decomposition.

(3) Extensive experiments demonstrate that DecoKAN achieves state-of-the-art long-term forecasting performance on benchmark and cryptocurrency datasets while offering robust interpretability through symbolic explanations.

\section{Related Work}

\subsection{Deep Learning Models for Time Series Forecasting}
Deep learning methods have become central to time series forecasting as data availability and computational capacity increase \cite{rojat2021explainable}. Transformer architectures, capable of modeling long-range dependencies, remain widely used. Representative models include Informer \cite{zhou2021informer}, Autoformer \cite{wu2021autoformer}, FEDformer \cite{zhou2022fedformer}, and Crossformer \cite{zhang2023crossformer}, while recent variants such as PatchTST \cite{nie2023time} and iTransformer \cite{liu2023itransformer} enhance local semantic capture with improved efficiency. Some approaches also explore probabilistic forecasting \cite{salinas2020deepar} and generative pre-training for time series \cite{cao2024tempo}. Despite these advances, their internal reasoning remains difficult to interpret \cite{vaswani2017attention},\cite{ kong2025deep, wang2024deep}.
More recent work explores MLP-based architectures as lightweight yet competitive alternatives. Zeng et al. \cite{zeng2023transformers} showed that simple linear models like DLinear can rival Transformer performance on standard benchmarks. Building on this insight, methods such as TimeMixer and WPMixer integrate decomposition mechanisms to separate trend and seasonal components or to perform multi-level wavelet analysis \cite{wang2023timemixer, murad2025wpmixer}. Frequency-domain approaches \cite{yi2024frequency, woo2022etsformer} and long-term forecasting methods \cite{das2023longterm} further enhance modeling capabilities. Other variants, including TimeXer and TimesNet, adapt MLP-Mixer and 2D-kernel designs for temporal data \cite{chen2023tsmixer, wu2023timesnet, tolstikhin2021mlpmixer}. Additionally, graph-based approaches like TimeFilter \cite{hu2025timefilter} employ spatial-temporal graph filtration to adaptively model dependencies while filtering out irrelevant correlations.  

Recent advances in deep learning have significantly improved time series forecasting performance, while decomposition-based approaches such as wavelet analysis have demonstrated strong potential for modeling and decoupling composite signals. 
Yet, even models employing such decomposition techniques often remain opaque, limiting their interpretability and undermining user trust in their predictions.

\subsection{Cryptocurrency Market Time Series Forecasting} 
Cryptocurrency market data are characterized by volatility, non-stationarity, and the complex composite nature of long-term trends and high-frequency oscillations, all of which complicate accurate forecasting. Earlier studies relied on statistical models such as ARIMA and VAR or on classical machine learning methods including SVMs and Random Forests \cite{box2015time}. While interpretable, these approaches fail to capture nonlinear dynamics and sudden structural shifts typical of crypto markets.
Deep learning models, from LSTMs to Transformers and recent adaptive distillation frameworks, have since been adopted for cryptocurrency forecasting \cite{ liu2024role,wu2025multi}. Hybrid architectures combining LSTM with Transformers have also emerged to leverage strengths of both approaches. However, they still struggle with the coexistence of slow, long-term trends and rapid, high-frequency fluctuations. Although wavelet decomposition effectively decouples signal components in general time series analysis, its use in digital asset analysis remains limited \cite{gencay2001introduction, mousa2025forecasting, kikuchi2024wavelet}. 
Furthermore, while recent studies have explored hybrid approaches combining machine learning with Large Language Models (LLMs) to balance performance and interpretability, existing methods often sacrifice decoupling capability for post-hoc explanations rather than achieving intrinsic transparency \cite{yu2023temporal, liu2023financial}.

Consequently, a critical research gap persists in these high-stakes financial domains: the lack of a unified methodology that can simultaneously disentangle composite socio-economic signals for high accuracy and provide the intrinsic, auditable interpretability required for trustworthy decision-making in complex social systems.

\subsection{Interpretable Deep Learning and Kolmogorov-Arnold Networks (KANs)} 
Efforts to make AI systems more transparent have driven research on interpretable deep learning \cite{brigo2021interpretability}. Two major approaches dominate the field: post-hoc explanation tools such as LIME \cite{ribeiro2016why} and SHAP \cite{lundberg2017unified}, which approximate reasoning after training, and intrinsically interpretable architectures designed for transparency from the outset \cite{oreshkin2020nbeats, lim2021temporal}. 
Among these, Kolmogorov–Arnold Networks (KANs) represent a promising development \cite{liu2025kan, liu2024kan}. Instead of fixed activation functions, KANs employ learnable spline functions along their connections, enabling both strong approximation guarantees and symbolic representation.
Interpretability in KANs is achieved through a structured process of sparsification, pruning, and symbolization that converts trained models into explicit mathematical expressions. This approach unites theoretical expressiveness with explicit transparency, forming a novel path toward explainable AI. KANs have shown promise across diverse domains, including recommendation systems \cite{park2024cfkan}, medical image analysis \cite{li2024ukan}, and various time series applications \cite{xu2024kolmogorov}. However, their application to complex, composite, and non-stationary time series, such as those in volatile financial markets, is still at an early stage \cite{koenig2024kanodes, vacarubio2024kan}.
While concurrent work, such as Wav-KAN \cite{bozorgasl2024wav}, modifies the KAN architecture itself by incorporating wavelet functions as activations, and other recent studies explore KAN for time series classification \cite{dong2024kolmogorov} and domain-specific forecasting tasks \cite{gao2025revolutionary, chen2024bilstmkan}, the effective combination of standard KANs' symbolic transparency with hierarchical signal decomposition via DWT, specifically tailored for cryptocurrency's distinctive temporal dynamics, remains largely unaddressed.

In essence, KANs bridge numerical learning and symbolic understanding. Their potential for interpretable forecasting is evident, yet their capacity to manage composite nature of market data has not been fully examined.

\section{Proposed Method}
Let $\mathcal{X} = \{x_t\}_{t=1}^N$ represent the historical multivariate time series data from a cryptocurrency market, 
where $N$ denotes the total length of the series. Each $x_{t}\in\mathbb{R}^{1\times C}$ constitutes an observation vector at time $t$, encompassing $C$ variates (e.g., transaction flows, market volatility, or token prices).
Given the historical sequence over a look-back window of length $L$, denoted as $X_{L}=\{x_{t-L+1},...,x_t\}$, our objective is to learn a mapping $F$ to forecast the future sequence $X_{T}=\{x_{t+1},...,x_{t+T}\}$ over a prediction horizon of length $T$. This forecast, representing the anticipated evolution of these key market indicators, is crucial for applications ranging from financial decision-making to quantitative investment analysis. The forecasting task can be formally expressed as:
\begin{equation}
X_T = F(X_L)
\end{equation}
where $X_T$ is the predicted future sequence.

\begin{figure*}[!t]
\centering
\includegraphics[width=1\textwidth]{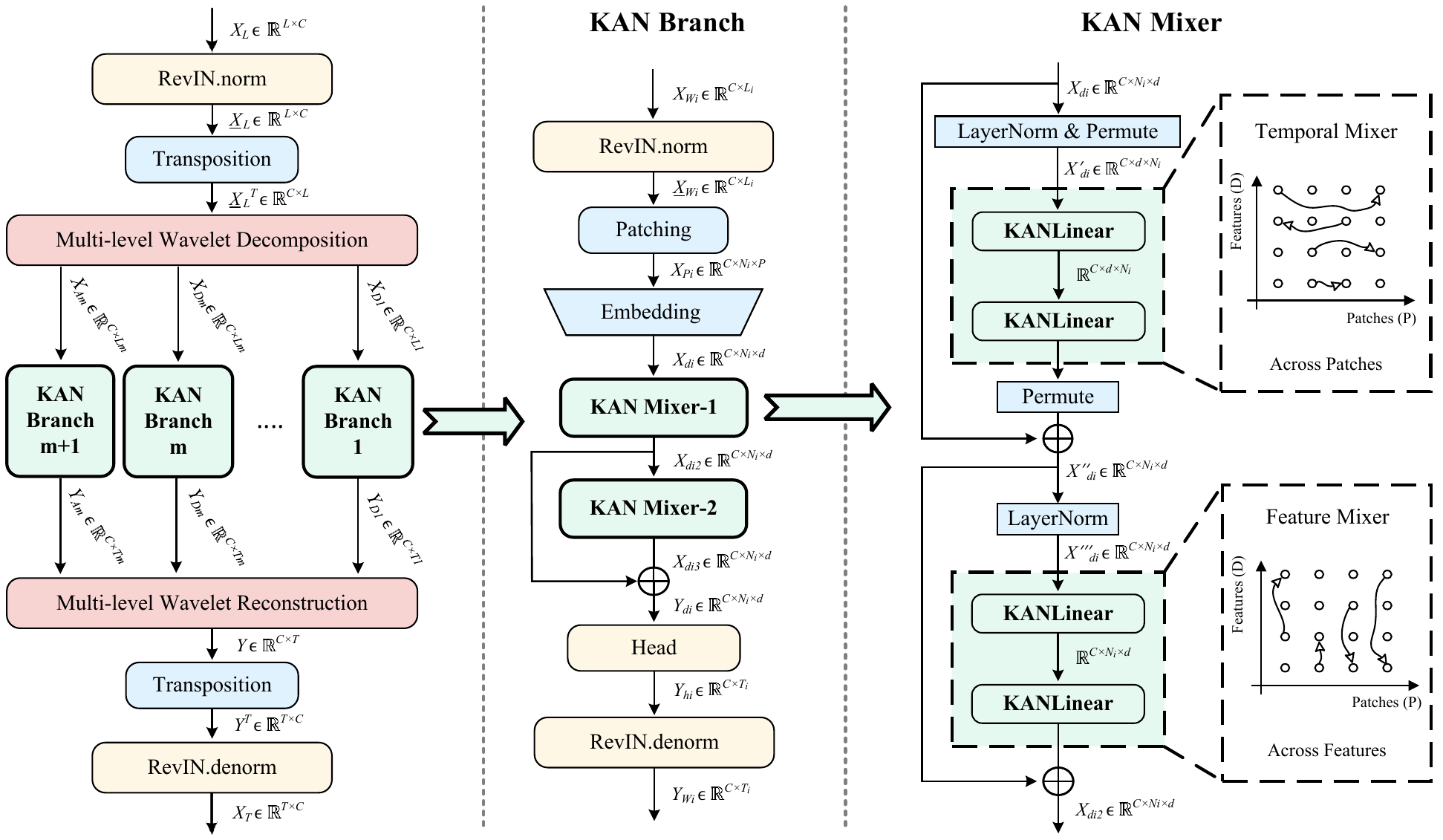}
\caption{Architecture overview of the proposed DecoKAN model. The model follows a decompose-mix-reconstruct paradigm: (1) Input time series is decomposed into multiple wavelet coefficient series using multi-level DWT; (2) Each coefficient series is processed by dedicated KAN Resolution Branches containing DecoKAN Mixer blocks; (3) Predicted coefficients are reconstructed back to time-domain via inverse wavelet transform.}
\label{fig:wavekan_architecture}
\end{figure*}

\subsection{Model Architecture}

To effectively model the disparate dynamics inherent in cryptocurrency markets, the architecture of our proposed model, DecoKAN, follows the principled decompose-mix-reconstruct paradigm. Fig. 2 provides an overview of the end-to-end pipeline. The core innovation of DecoKAN lies in replacing the conventional opaque MLP-based mixers with more expressive and interpretable Kolmogorov-Arnold Networks (KANs), performing all core mixing operations within the wavelet domain.

DecoKAN begins by robustly normalizing the input time series and then decomposing it into multiple approximation and detail coefficient series via a multi-level wavelet transform. This decomposition allows for feature extraction tailored to these decoupled components. As shown in Fig. 2 (left panel), each resulting coefficient series is then processed independently by a dedicated KAN Resolution Branch. 
This branch structure facilitates the parallel and independent processing of the approximation and detail components, minimizing spectral interference and enabling frequency-specific pattern learning.
Within each branch (Fig. 2, middle panel), the coefficient series undergoes patching and embedding before being processed sequentially by two DecoKAN Mixer blocks. These mixers model complex temporal and feature-wise dependencies. Finally, the predicted coefficients from all branches are synthesized back into a coherent time-domain signal via an Inverse Wavelet Transform (IDWT), followed by denormalization to produce the final forecast.
\subsection{Instance Normalization}

Given the extreme volatility and non-stationarity characteristic of cryptocurrency data, mitigating distribution shift is crucial for model generalization. We employ Reversible Instance Normalization (RevIN) for this purpose, applied without learnable affine parameters consistent with our implementation \cite{kim2022reversible}. As depicted in Fig. 2, RevIN normalizes the input sequence $X_L$ by subtracting the channel-wise mean and dividing by the standard deviation before the decomposition stage, storing these statistics. Our implementation includes adding an epsilon (1e-6) to the variance and clamping the standard deviation (min 1e-4) for numerical stability. 
The inverse operation (RevIN.denorm) is applied after reconstruction. Normalization is also applied within each resolution branch. The subsequent transposition ${\underline{X}_{L}}^{\top} \in \mathbb{R}^{C \times L}$ aligns the data tensor with the channel-first convention typically expected by DWT implementations.

\subsection{Multi-level Wavelet Decomposition}

To decompose the composite signals inherent in cryptocurrency market data—separating underlying low-frequency trends from high-frequency noise and volatility—we adopt the multi-level Discrete Wavelet Transform (DWT) approach, where the normalized, transposed time series $\underline{X}_L$ is decomposed by iteratively applying high-pass and low-pass filters derived from the chosen wavelet basis.

We select the Daubechies 4 ('db4') wavelet, denoted by $\psi$, motivated by its balance between smoothness and compact support, suitable for capturing both transient and trend components in in non-stationary time series. The number of decomposition levels, $m$, is a hyperparameter (defaulting to 1), typically chosen based on the input length $L$ (e.g., $m \approx \lfloor \log_2 L \rfloor - k$, for a small $k$) and validated empirically to balance resolution and computational cost. Symmetric padding is used to handle edge effects during the DWT process. The decomposition yields $m+1$ coefficient series:
\begin{equation}
[X_{A_m},X_{D_m},X_{D_{m-1}},...,X_{D_1}]=\text{DWT}(\underline{X}_{L},\psi,m)
\end{equation}
\noindent
where $X_{A_m} \in \mathbb{R}^{C \times L_m}$ is the approximation coefficient series at level $m$, and $X_{D_m} \in \mathbb{R}^{C \times L_m}$, ..., $X_{D_1} \in \mathbb{R}^{C \times L_1}$ are the detail coefficient series at their respective levels. Let $X_{w_i}$ denote one of these $m+1$ coefficient series (e.g., $X_{w_i} = X_{D_1}$ with input length $L_i = L_1$). Each $X_{w_i} \in \mathbb{R}^{C \times L_i}$ is processed by a dedicated KAN Resolution Branch.

\subsection{KAN Branch}

This section details the core processing unit of our model. Each KAN Resolution Branch first normalizes its input $X_{w_i} \in \mathbb{R}^{C \times L_i}$ using an internal RevIN.norm, producing $\underline{X}_{w_i} \in \mathbb{R}^{C \times L_i}$. This normalized series then undergoes patching and embedding. The series is padded and divided into overlapping patches, 
creating $X_{P_i} \in \mathbb{R}^{C \times N_i \times P}$, where $N_i$ denotes the number of patches and $P$ denotes the patch size (with stride $S$). These are projected to dimension $d$ via a shared linear layer:
\begin{equation}
X_{d_i} = \text{Embedding}(X_{P_i}) \in \mathbb{R}^{C \times N_i \times d}
\end{equation}
This tensor $X_{d_i}$ serves as the input to the first of two sequential KAN Mixer blocks and also as the starting point for a branch-level residual connection (see Fig. 2, middle panel). The data flows sequentially through the mixers:
\begin{align}
X_{d_i2} &= \text{KAN Mixer-1}(X_{d_i}) \\
X_{d_i3} &= \text{KAN Mixer-2}(X_{d_i2})
\end{align}
The final mixed representation $Y_{d_i}$ combines the output of the second mixer with the original embedded input via this branch-level residual connection:
\begin{equation}
Y_{d_i} = X_{d_i} \oplus X_{d_i3}
\end{equation}
The tensor $Y_{d_i} \in \mathbb{R}^{C \times N_i \times d}$ is the tensor finally passed to the Head module.

\textbf{KAN Mixer:} Each KAN Mixer block (Fig. 2, right panel) models temporal and feature interactions sequentially using KAN-based layers. It consists of two sub-modules with residual connections. The process for KAN Mixer-1 is as follows:

\begin{enumerate}[leftmargin=12pt, itemindent=0pt, labelsep=0.5em]
\item \textbf{Temporal KAN Mixer}: Learns relationships along the temporal (patch, $N_i$) axis. The input $X_{d_i}$ is normalized ($\mathcal{N}$), permuted ($\mathcal{P}_{N \leftrightarrow d}$, shape $C \times d \times N_i$), processed by the temporal KAN ($KAN_p$, which operates on the $N_i$ dimension), and permuted back ($\mathcal{P}_{d \leftrightarrow N}$). This output is added to $X_{d_i}$ via a residual connection (with Dropout, $\mathcal{D}$):
\begin{align}
Y' &= \mathcal{P}_{d \leftrightarrow N}( KAN_p( \mathcal{P}_{N \leftrightarrow d}( \mathcal{N}(X_{d_i}) ) ) ) \\
X''_{d_i} &= X_{d_i} + \mathcal{D}(Y')
\end{align}

\item \textbf{Feature KAN Mixer}: Captures interactions across the feature (embedding, $d$) dimension. The intermediate tensor $X''_{d_i}$ is normalized, processed by the feature KAN ($KAN_e$, which operates on the $d$ dimension), and added to $X''_{d_i}$ via a second residual connection:
\begin{align}
Y'' &= KAN_e( \mathcal{N}(X''_{d_i}) ) \\
X_{d_i2} &= X''_{d_i} + \mathcal{D}(Y'')
\end{align}
\end{enumerate}
The tensor $X_{d_i2}$ is the final output of KAN Mixer-1. KAN Mixer-2 follows the same process as KAN Mixer-1.

The advantage of these KAN-based mixers in financial modeling stems from their core KANLinear layer.
Each connection uses a learnable univariate activation $\phi(x)$ 
(default $grid size=5$ and $spline order=3$),
parameterized via B-splines over a grid, typically added to a base function $b(x)$ (SiLU in our implementation):
\begin{equation}
\phi(x) = w_b b(x) + w_s \sum_{j=0}^{G+k-1} c_j B_j(x) 
\end{equation}
where $B_j(x)$ are B-spline basis functions, and $w_b, w_s, c_j$ are learnable.

Furthermore, to enhance interpretability and encourage sparsity, each KANLinear layer incorporates a regularization loss based on:
\begin{equation}
L_{\rm reg}=\lambda_1 \sum_{\text{edges}} |\phi|_{1} + \lambda_2 \sum_{\text{edges}} S(\phi)
\end{equation}

This composite loss function is crucial for learning sparse and interpretable representations. It consists of two key components, balanced by internal coefficients $\lambda_1$ and $\lambda_2$ (typically set to 1.0), and weighted by an overall regularization strength hyperparameter $\gamma$ (e.g., $\gamma=1e-5$) in the total loss $L_{\rm total} = L_{\rm forecast} + \gamma L_{\rm reg}$, where $L_{\rm forecast}$ denotes the Mean Squared Error (MSE) between the predicted and ground-truth values, and $\gamma$ is a hyperparameter balancing the two terms.

\begin{itemize}[leftmargin=12pt, itemindent=0pt, labelsep=0.5em]
\item An L1-based term $|\phi|_{avg}$ representing the average magnitude of the spline component, encouraging function sparsity.
\item An entropy term $S(\phi)$ based on the normalized magnitudes ($p_j$) of the spline coefficients ($c_j$), $S(\phi) = -\sum_j p_j \log p_j$ (where $p_j = |c_j| / \sum_k |c_k|$), encouraging structural sparsity at the neuron level.
\end{itemize}

This dual-penalty regularization, controlled overall by $\gamma$, guides the model towards parsimonious and interpretable solutions during training.

\subsection{Multi-level Wavelet Reconstruction}

After the branch-level residual connection, the resulting tensor $Y_{d_i} \in \mathbb{R}^{C \times N_i \times d}$ is fed into the Head module. This module first flattens the patch and embedding dimensions using nn.Flatten, then uses a final nn.Linear layer to project the representation to the desired prediction length $T_i$ for that specific coefficient series:
\begin{align}
Y_{f_i} &= \text{Flatten}(Y_{d_i}) \in \mathbb{R}^{C \times (N_i \cdot d)} \\
Y_{h_i} &= \text{Linear}(Y_{f_i}) \in \mathbb{R}^{C \times T_i}
\end{align}
These outputs $Y_{h_i}$ correspond to the predicted coefficients we defined earlier: $Y_{A_m} \in \mathbb{R}^{C \times T_m}$ and $Y_{D_m} \in \mathbb{R}^{C \times T_m}$, ..., $Y_{D_1} \in \mathbb{R}^{C \times T_1}$.

Finally, the Reconstruction stage synthesizes these predicted coefficients back into the time domain. The predicted approximation ($Y_{A_m}$) and detail ($Y_{D_m}, ..., Y_{D_1}$) coefficient series from all resolution branches are combined using the inverse multi-level wavelet transform (IDWT):
\begin{equation}
  Y = \text{IDWT}(Y_{A_m}, Y_{D_m}, ..., Y_{D_1}) \in \mathbb{R}^{C \times T}
\end{equation}

The resulting time series $Y$ represents the forecast in the normalized, channel-first space. This tensor is then transposed to $Y^\top\in \mathbb{R}^{T \times C}$ and transformed using the stored RevIN parameters (denormalization) to obtain the final prediction sequence $X_T \in \mathbb{R}^{T \times C}$ in the original data space.

\begin{algorithm}[H]
\caption{DecoKAN Framework: Training and Interpretation}
\label{alg:decokan_full}
\begin{algorithmic}
\REQUIRE Multivariate time series $\mathcal{X}$, Hyperparameters ($L, T, m, \gamma, \tau$)
\ENSURE Trained Model $\theta^*$, Symbolic Formulas $\mathcal{F}$
\STATE \textbf{Phase 1: Structure-Aware Training}
\FOR{epoch $= 1$ to $E$}
    \FOR{batch $(X_{L}, X_{\rm true})$ in DataLoader}
        \STATE \textit{// 1. Hierarchical Decomposition}
        \STATE $\tilde{X}_{L} \leftarrow \text{Normalize}(X_{L})$
        \STATE $\mathcal{C} = [X_{A_m}, X_{D_m}, \dots, X_{D_1}] \leftarrow \text{DWT}(\tilde{X}_{L}, m)$
        \STATE \textit{// 2. Parallel Resolution Branch Processing}
        \STATE $Y_{\rm coeffs} \leftarrow []$; $L_{\rm reg} \leftarrow 0$
        \FOR{each component $c_i$ in $\mathcal{C}$}
            \STATE $Z_i \leftarrow \text{Embed}(c_i)$  \textit{// Implicit Patching \& Normalization}
            \STATE $H_i \leftarrow \text{KAN\_Block}(Z_i)$ \textit{// Capture non-linear patterns via Splines}
            \STATE $\hat{y}_i \leftarrow \text{ProjectionHead}(H_i)$
            \STATE \textbf{// Sparsification via Regularization}
            \STATE $L_{\rm reg} \leftarrow L_{\rm reg} + \text{SparsityLoss}(\phi_i)$
            \STATE $Y_{\rm coeffs}.\text{append}(\hat{y}_i)$
        \ENDFOR
        \STATE \textit{// 3. Reconstruction \& Optimization}
        \STATE $\hat{X}_{T} \leftarrow \text{Denormalize}(\text{IDWT}(Y_{\rm coeffs}))$
        \STATE $L_{\rm total} \leftarrow \text{MSE}(\hat{X}_{T}, X_{\rm true}) + \gamma \cdot L_{\rm reg}$
        \STATE Update $\theta$ by minimizing $L_{\rm total}$
    \ENDFOR
\ENDFOR
\STATE \textbf{Phase 2: Interpretability (Post-Training)}
\STATE Initialize $\mathcal{F} \leftarrow \emptyset$
\FOR{each KAN layer $l$ in optimized $\theta^*$}
    \STATE \textbf{Pruning:} Mask connections where $||\phi_l||_2 < \tau$
    \STATE \textbf{Symbolification:} Extract $f_l \approx \phi_l$ maximizing $R^2$ and update $\mathcal{F} \leftarrow \mathcal{F} \cup \{f_l\}$
\ENDFOR
\RETURN $\theta^*$ and $\mathcal{F}$
\end{algorithmic}
\end{algorithm}

\section{Experiments and Analysis}
To validate the effectiveness, interpretability, and robustness of our proposed DecoKAN framework, we conducted extensive experiments across various benchmarks. 
\subsection{Experimental Setup}

\textbf{Datasets.} Our evaluation employs a diverse set of time series datasets, encompassing both widely-adopted general benchmarks and high-volatility, real-world cryptocurrency market data, as summarized in Table I.

(1) \textbf{General Benchmarks:} For long-term forecasting, we utilize the widely-adopted ETT datasets (ETTh1, ETTh2, ETTm1, ETTm2). These datasets, collected from electricity transformers, are commonly used to assess time series forecasting models due to their diverse temporal patterns and varying frequencies.

(2) \textbf{Cryptocurrency Benchmarks:} To rigorously test DecoKAN's capability in its primary application domain, we introduce three real-world cryptocurrency datasets: Bitcoin (BTC), Ethereum (ETH), and Monero (XMR), sourced from the Community Network Data provided by Coin Metrics \cite{coinmetrics}. These datasets comprise comprehensive daily multivariate time series spanning diverse dimensions of the digital asset market. Key categories include market dynamics (e.g., price, market cap, volatility), on-chain activity (e.g., transaction counts, fees), and network security (e.g., hashrate, difficulty). Collectively, this rich feature set captures the complex interplay between financial sentiment, fundamental network utility, and market fundamentals, characterized by extreme volatility and complex non-stationarity.

\begin{table}[H]
\renewcommand{\arraystretch}{1.2}
\centering
\caption{Dataset Statistics Summary}
\label{tab:dataset_statistics}
\begin{tabular}{c|c|c|c}
\toprule
\textbf{Dataset} & \textbf{Variates} & \textbf{Dataset Size} & \textbf{Freq.} \\
\midrule
ETTh1, ETTh2 & 7 & (8545, 2881, 2881) & Hourly \\
ETTm1, ETTm2 & 7 & (34465, 11521, 11521) & 15 min \\
BTC & 147 & (6099, 4269, 1219) & Daily \\
ETH & 146 & (3700, 2590, 740) & Daily \\
XMR & 48 & (4168, 2917, 833) & Daily \\
\bottomrule
\end{tabular}
\end{table}

\begin{table*}[!b]
\renewcommand{\arraystretch}{0.9}
\setcounter{table}{2}
\caption{Experimental Results on Long-term Time Series Forecasting. Results are reported as MSE/MAE. Best results are in \textbf{bold}, second-best are \underline{underlined}. The length of the look-back window is a hyperparameter.}
\label{tab:main_results}
\centering
\footnotesize
\setlength{\tabcolsep}{1.2mm}
\begin{tabular}{lccccccccccccccccccc}
\toprule
\multicolumn{2}{c}{\textbf{Models}} & \multicolumn{2}{c}{\textbf{DecoKAN}} & \multicolumn{2}{c}{\textbf{WPMixer}} & \multicolumn{2}{c}{\textbf{TimeFilter}} & \multicolumn{2}{c}{\textbf{TimeMixer}} & \multicolumn{2}{c}{\textbf{TimeXer}} & \multicolumn{2}{c}{\textbf{TimesNet}} & \multicolumn{2}{c}{\textbf{PatchTST}} & \multicolumn{2}{c}{\textbf{DLinear}} & \multicolumn{2}{c}{\textbf{Crossformer}}
\\
& & \multicolumn{2}{c}{\textbf{(Ours)}} & \multicolumn{2}{c}{\textit{AAAI'25}} & \multicolumn{2}{c}{\textit{ICML'25}} & \multicolumn{2}{c}{\textit{ICLR'24}} & \multicolumn{2}{c}{\textit{NeurIPS'24}} & \multicolumn{2}{c}{\textit{ICLR'23}} & \multicolumn{2}{c}{\textit{ICLR'23}} & \multicolumn{2}{c}{\textit{AAAI'23}} & \multicolumn{2}{c}{\textit{ICLR'23}} \\
\midrule
\multicolumn{2}{c}{Metric}& MSE&MAE & MSE&MAE & MSE&MAE & MSE&MAE & MSE&MAE & MSE&MAE & MSE&MAE & MSE&MAE & MSE&MAE \\
\midrule
\multirow{5}{*}{ETTh1} 
& 96 & \textbf{0.370} & \underline{0.399} & \underline{0.371} & \underline{0.399} & 0.389 & \underline{0.399} & 0.375 & \textbf{0.397} & 0.383 & 0.403 & 0.407 & 0.425 & 0.382 & 0.400 & 0.396 & 0.411 & 0.451 & 0.455 \\
& 192 & \textbf{0.406} & \textbf{0.418} & 0.432 & 0.440 & 0.440 & \underline{0.427} & \underline{0.429} & \underline{0.427} & 0.442 & 0.438 & 0.465 & 0.460 & 0.430 & 0.433 & 0.447 & 0.443 & 0.477 & 0.476 \\
& 336 & \textbf{0.439} & \textbf{0.444} & \underline{0.441} & 0.450 & 0.478 & \underline{0.445} & 0.481 & 0.451 & 0.483 & 0.452 & 0.496 & 0.471 & 0.472 & 0.460 & 0.496 & 0.473 & 0.572 & 0.530 \\
& 720 & \textbf{0.436} & \textbf{0.460} & \underline{0.489} & 0.486 & 0.507 & 0.482 & 0.490 & \underline{0.476} & 0.520 & 0.494 & 0.513 & 0.495 & 0.513 & 0.501 & 0.510 & 0.508 & 0.913 & 0.743 \\
\arrayrulecolor{gray}\cmidrule(lr){2-20}\arrayrulecolor{black}
& Avg & \textbf{0.413} & \textbf{0.430} & \underline{0.433} & 0.444 & 0.454 & \underline{0.438} & 0.444 & \underline{0.438} & 0.457 & 0.447 & 0.471 & 0.463 & 0.449 & 0.448 & 0.462 & 0.459 & 0.603 & 0.551\\
\cmidrule{1-20}
\multirow{5}{*}{ETTh2} 
& 96 & \textbf{0.278} & \underline{0.341} & \underline{0.281} & 0.348 & 0.287 & \textbf{0.337} & 0.291 & \underline{0.341} & 0.283 & \textbf{0.337} & 0.345 & 0.381 & 0.304 & 0.353 & 0.350 & 0.402 & 0.635 & 0.555 \\
& 192 & \textbf{0.365} & 0.396 & \underline{0.368} & 0.402 & 0.376 & \underline{0.394} & 0.373 & \textbf{0.393} & \underline{0.368} & \textbf{0.393} & 0.423 & 0.417 & 0.379 & 0.400 & 0.463 & 0.469 & 0.578 & 0.548 \\
& 336 & \textbf{0.373} & \textbf{0.404} & \underline{0.381} & \underline{0.420} & 0.430 & 0.439 & 0.437 & 0.433 & 0.434 & 0.438 & 0.444 & 0.453 & 0.400 & 0.452 & 0.573 & 0.533 & 0.896 & 0.669 \\
& 720 & \underline{0.402} & \underline{0.439} & \textbf{0.395} & \textbf{0.437} & 0.447 & 0.458 & 0.437 & 0.448 & 0.445 & 0.454 & 0.455 & 0.465 & 0.456 & 0.464 & 0.839 & 0.661 & 1.097 & 0.757 \\
\arrayrulecolor{gray}\cmidrule(lr){2-20}\arrayrulecolor{black}
& Avg & \textbf{0.354} & \textbf{0.395} & \underline{0.357} & \underline{0.402} & 0.385 & 0.407 & 0.385 & 0.404 & 0.383 & 0.406 & 0.417 & 0.429 & 0.385 & 0.417 & 0.556 & 0.516 & 0.802 & 0.632 \\
\cmidrule{1-20}
\multirow{5}{*}{ETTm1} 
& 96 & \underline{0.303} & \underline{0.354} & \textbf{0.300} & \textbf{0.349} & 0.320 & 0.359 & 0.318 & 0.358 & 0.322 & 0.359 & 0.328 & 0.369 & 0.324 & 0.364 & 0.345 & 0.372 & 0.403 & 0.412 \\
& 192 & \underline{0.337} & \underline{0.374} & \textbf{0.336} & \textbf{0.371} & 0.362 & 0.381 & 0.360 & 0.380 & 0.365 & 0.385 & 0.411 & 0.407 & 0.370 & 0.390 & 0.382 & 0.390 & 0.477 & 0.458 \\
& 336 & \textbf{0.369} & \underline{0.399} & \underline{0.372} & \textbf{0.390} & 0.391 & 0.403 & 0.385 & 0.400 & 0.401 & 0.409 & 0.423 & 0.426 & 0.398 & 0.409 & 0.415 & 0.414 & 0.474 & 0.472 \\
& 720 & \textbf{0.421} & \textbf{0.417} & \underline{0.435} & \underline{0.423} & 0.460 & 0.438 & 0.454 & 0.442 & 0.453 & 0.441 & 0.493 & 0.463 & 0.464 & 0.447 & 0.472 & 0.450 & 0.532 & 0.503 \\
\arrayrulecolor{gray}\cmidrule(lr){2-20}\arrayrulecolor{black}
& Avg & \textbf{0.358} & \underline{0.386} & \underline{0.361} & \textbf{0.383} & 0.383 & 0.395 & 0.379 & 0.395 & 0.385 & 0.399 & 0.414 & 0.416 & 0.389 & 0.402 & 0.403 & 0.407 & 0.472 & 0.461 \\
\cmidrule{1-20}
\multirow{5}{*}{ETTm2} 
& 96 & \underline{0.170} & 0.261 & \textbf{0.168} & \textbf{0.257} & 0.173 & 0.259 & 0.176 & \underline{0.258} & 0.171 & \textbf{0.257} & 0.185 & 0.264 & 0.180 & 0.265 & 0.194 & 0.293 & 0.285 & 0.370 \\
& 192 & \textbf{0.228} & \textbf{0.298} & \underline{0.232} & 0.305 & 0.237 & \underline{0.300} & 0.242 & 0.302 & 0.240 & 0.301 & 0.256 & 0.310 & 0.247 & 0.311 & 0.284 & 0.361 & 0.388 & 0.438 \\
& 336 & \textbf{0.276} & \textbf{0.330} & \underline{0.277} & \underline{0.331} & 0.296 & 0.338 & 0.305 & 0.346 & 0.297 & 0.339 & 0.314 & 0.345 & 0.314 & 0.352 & 0.373 & 0.421 & 0.613 & 0.541 \\
& 720 & \textbf{0.381} & \textbf{0.396} & 0.413 & 0.411 & 0.397 & \underline{0.398} & 0.396 & 0.400 & \underline{0.394} & \textbf{0.396} & 0.421 & 0.409 & 0.421 & 0.415 & 0.538 & 0.515 & 1.356 & 0.785 \\
\arrayrulecolor{gray}\cmidrule(lr){2-20}\arrayrulecolor{black}
& Avg & \textbf{0.264} & \textbf{0.321} & \underline{0.272} & 0.326 & 0.276 & 0.324 & 0.280 & 0.327 & 0.275 & \underline{0.323} & 0.294 & 0.332 & 0.290 & 0.336 & 0.347 & 0.398 & 0.661 & 0.534 \\
\toprule
\multirow{5}{*}{BTC} 
& 24 & \underline{0.108} & \textbf{0.096} & 0.134 & 0.133 & 0.110 & 0.106 & 0.146 & 0.112 & \underline{0.108} & 0.101 & 0.138 & 0.135 & \textbf{0.107} & \underline{0.098} & 0.109 & 0.115 & 0.145 & 0.167 \\
& 48 & 0.120 & \textbf{0.115} & 0.153 & 0.152 & 0.121 & \underline{0.116} & 0.131 & 0.123 & \underline{0.119} & 0.119 & 0.141 & 0.146 & \textbf{0.118} & \underline{0.116} & 0.123 & 0.145 & 0.186 & 0.210 \\
& 96 & \textbf{0.142} & \textbf{0.138} & 0.163 & 0.168 & 0.158 & \underline{0.142} & 0.160 & 0.154 & \underline{0.143} & 0.145 & 0.166 & 0.168 & 0.144 & 0.146 & 0.156 & 0.188 & 0.221 & 0.245 \\
& 168 & \underline{0.174} & \underline{0.172} & 0.190 & 0.192 & 0.193 & 0.175 & \textbf{0.172} & \textbf{0.166} & 0.178 & 0.177 & 0.210 & 0.213 & 0.181 & 0.177 & 0.198 & 0.227 & 0.293 & 0.309 \\
\arrayrulecolor{gray}\cmidrule(lr){2-20}\arrayrulecolor{black}
& Avg & \textbf{0.136} & \textbf{0.130} & 0.160 & 0.161 & 0.146 & 0.135 & 0.152 & 0.139 & \underline{0.137} & 0.135 & 0.164 & 0.166 & \underline{0.137} & \underline{0.134} & 0.146 & 0.169 & 0.211 & 0.233 \\
\cmidrule{1-20}
\multirow{5}{*}{ETH} 
& 24 & \textbf{0.061} & \textbf{0.108} & 0.110 & 0.174 & 0.078 & 0.121 & 0.072 & 0.114 & 0.064 & 0.114 & 0.109 & 0.174 & \underline{0.062} & \underline{0.112} & \underline{0.062} & 0.128 & 0.088 & 0.171 \\
& 48 & \textbf{0.084} & \textbf{0.136} & 0.133 & 0.195 & 0.097 & 0.144 & 0.099 & 0.145 & 0.091 & \underline{0.143} & 0.129 & 0.193 & \underline{0.088} & \underline{0.143} & 0.089 & 0.169 & 0.107 & 0.190 \\
& 96 & \textbf{0.110} & \textbf{0.169} & 0.153 & 0.220 & 0.128 & \underline{0.172} & 0.136 & 0.188 & 0.115 & 0.173 & 0.158 & 0.223 & \underline{0.114} & 0.179 & 0.126 & 0.229 & 0.154 & 0.260 \\
& 168 & \textbf{0.144} & 0.205 & 0.168 & 0.242 & 0.167 & \underline{0.203} & 0.168 & 0.219 & 0.177 & 0.223 & 0.205 & 0.261 & \underline{0.147} & \textbf{0.198} & 0.181 & 0.304 & 0.198 & 0.317 \\
\arrayrulecolor{gray}\cmidrule(lr){2-20}\arrayrulecolor{black}
& Avg & \textbf{0.100} & \textbf{0.155} & 0.141 & 0.208 & 0.118 & \underline{0.160} & 0.119 & 0.166 & 0.112 & 0.163 & 0.150 & 0.213 & \underline{0.103} & 0.162 & 0.115 & 0.208 & 0.137 & 0.234 \\
\cmidrule{1-20}
\multirow{5}{*}{XMR} 
& 24 & \underline{0.142} & \underline{0.112} & 0.189 & 0.153 & 0.165 & 0.113 & 0.143 & \textbf{0.109} & 0.144 & 0.114 & 0.210 & 0.158 & 0.284 & 0.143 & \textbf{0.141} & 0.117 & 0.157 & 0.165 \\
& 48 & \textbf{0.185} & \underline{0.139} & 0.239 & 0.177 & 0.190 & 0.140 & \underline{0.186} & \textbf{0.137} & 0.199 & 0.148 & 0.273 & 0.184 & 0.382 & 0.183 & 0.192 & 0.160 & 0.193 & 0.190 \\
& 96 & \textbf{0.239} & \underline{0.175} & 0.279 & 0.202 & 0.262 & \textbf{0.172} & 0.261 & 0.177 & \underline{0.253} & 0.178 & 0.297 & 0.210 & 0.576 & 0.236 & 0.254 & 0.211 & 0.241 & 0.214 \\
& 168 & \textbf{0.310} & \textbf{0.215} & 0.324 & 0.227 & 0.326 & \underline{0.216} & 0.336 & 0.224 & \underline{0.313} & 0.217 & 0.364 & 0.257 & 0.580 & 0.253 & 0.330 & 0.248 & 0.325 & 0.279 \\
\arrayrulecolor{gray}\cmidrule(lr){2-20}\arrayrulecolor{black}
& Avg & \textbf{0.219} & \textbf{0.160} & 0.258 & 0.190 & 0.235 & \textbf{0.160} & 0.231 & \underline{0.162} & \underline{0.227} & 0.164 & 0.286 & 0.202 & 0.455 & 0.204 & 0.229 & 0.184 & 0.229 & 0.212\\
\cmidrule{1-20}
\multicolumn{2}{c}{1stCount:} & 27 & 21 & 4&6 &0&3 & 1&5 & 0&4 & 0&0 & 2&1 & 1&0 & 0&0 \\
\bottomrule
\end{tabular}
\end{table*}

\textbf{Baselines.} We conduct a comprehensive comparison of DecoKAN against a broad spectrum of state-of-the-art deep forecasting models, representing different architectural backbones including MLP-based, Transformer-based, CNN-based and Graph-based approaches.

\noindent 
\textbf{MLP-based models:} 
\begin{itemize}
\item WPMixer \cite{murad2025wpmixer} employs multi-level wavelet decomposition with MLP mixers, processing each resolution branch independently.
\item TimeMixer \cite{wang2023timemixer} uses moving averages for seasonal-trend decomposition combined with multi-scale mixing.
\item DLinear \cite{zeng2023transformers} is a simple linear model shown to be remarkably effective, often used as a strong baseline.
\end{itemize}
\noindent 
\textbf{Transformer-based models:}
\begin{itemize}
  \item  TimeXer \cite{chen2023tsmixer} extends the MLP-Mixer paradigm incorporating Transformer-like elements for time series.
  \item  PatchTST \cite{nie2023time} applies patching to the input sequence before feeding it into a standard Transformer architecture.
  \item  Crossformer \cite{zhang2023crossformer} employs a two-stage attention mechanism to capture cross-time and cross-variable dependencies.
\end{itemize}
\noindent 
\textbf{CNN-based model:}
\begin{itemize}
  \item  TimesNet \cite{wu2023timesnet} transforms 1D time series into a 2D space based on periodicity and uses CNNs to capture variations.
\end{itemize}
\noindent 
\textbf{Graph-based model:}
\begin{itemize}
  \item  TimeFilter \cite{hu2025timefilter} employs a GNN-based framework with patch-specific spatial-temporal graph filtration to adaptively model fine-grained dependencies while filtering out irrelevant correlations.
\end{itemize}
\newpage
\textbf{Experimental Settings.} To ensure a fair and reproducible comparison, our experimental setup strictly adheres to the standards established by the Time Series Library (TSLib) \cite{wu2023timesnet,wang2024deep}. All baseline models are evaluated using their officially reported optimal parameters. To ensure reproducibility and transparency, the detailed hyperparameter configurations for both the cryptocurrency datasets and general benchmarks are summarized in Table II. For all forecasting tasks, we employ Mean Squared Error (MSE) and Mean Absolute Error (MAE) as primary evaluation metrics. All experiments were conducted on a single NVIDIA Tesla V100 GPU. 

\subsection{Performance Comparison with Baselines}

The comprehensive long-term forecasting results, comparing DecoKAN against state-of-the-art baselines across all datasets, are presented in Table III. Performance is evaluated using Mean Squared Error (MSE) and Mean Absolute Error (MAE), where lower values are better.
\begin{table}[H]
\renewcommand{\arraystretch}{1.2}
\centering
\setcounter{table}{1}
\caption{Detailed Hyperparameter Configurations for DecoKAN. Values include the candidate search spaces and ranges explored during optimization.}
\label{tab:hyperparams_full}
\resizebox{\columnwidth}{!}{%
\begin{tabular}{l|c|c}
\toprule
\textbf{Parameter} & \textbf{Crypto Benchmarks} & \textbf{ETT Benchmarks} \\
\midrule
Look-back Window ($L$) & \textbf{96}  & \textbf{512}  \\
Prediction Horizon ($T$) & $\{24, 48, 96, 168\}$ & $\{96, 192, 336, 720\}$ \\
\midrule
Wavelet Basis ($\psi$) & \texttt{db4} & \texttt{db2} / \texttt{db4} \\
Decomposition Level ($m$) & $1,2$ & $1,2,3$ \\
KAN Grid Size ($G$) & 5 & 5 \\
KAN Spline Order ($k$) & 3 & 3 \\
\midrule
Patch Size ($P$) & $8, 16$ & $16, 48$ \\
Patch Stride ($S$) & $4,8$ & $8, 24$ \\
Model Dimension ($d$) & $64, 128, 256$ & $64, 128$ \\
Temporal Factors & $2 \sim 6$ & $2 \sim 6$ \\
Dimension Factors & $2 \sim 6$ & $2 \sim 8$ \\
Dropout Rate & $0.05 \sim 0.3$ & $0.0 \sim 0.4$ \\
\midrule
Batch Size & $4,8,16$ & $32,64,128$ \\
Learning Rate & $1\text{e}^{-4} \sim 5\text{e}^{-4}$ & $1\text{e}^{-4} \sim 1\text{e}^{-3}$ \\
epochs & $30$ & $30,50$ \\
\bottomrule
\end{tabular}%
}
\end{table}

On the general ETT benchmark datasets, DecoKAN demonstrates strong performance. It achieves the best overall average MSE and MAE on the ETTh1 dataset, reducing average MSE by approximately 4.6\% compared to the next best model, WPMixer. On ETTh2 and ETTm2, DecoKAN also yields the lowest average MSE among all models considered. While WPMixer shows a marginally better average MAE on ETTm1 , DecoKAN maintains competitive accuracy across most prediction horizons, particularly for longer-term forecasts (e.g., $T=720$) on ETTh1 and ETTh2 , underscoring the robustness of the wavelet-KAN architecture.

Notably, this solid performance translates into leading results on the highly volatile cryptocurrency datasets (BTC, ETH, XMR), which represent our primary application domain. DecoKAN achieves the lowest average MSE and MAE across all horizons for all three cryptocurrency datasets compared to all baselines listed in Table III. 
Specifically, compared to the recent WPMixer baseline, DecoKAN reduces the average MSE by approximately 15.0\% on BTC, 29.1\% on ETH, and 15.1\% on XMR. 
Furthermore, even against the state-of-the-art graph-based model TimeFilter, DecoKAN maintains a distinct advantage, achieving an overall average MSE reduction of 9.6\% across these cryptocurrency datasets.
Notably compared to the closest competitors like PatchTST on BTC/ETH or TimeXer on XMR, DecoKAN consistently maintains an edge in average MSE.

This observed advantage in the target domain aligns with our design principles. We attribute this effective performance primarily to DecoKAN's architectural strengths: the multi-level wavelet decomposition effectively disentangles the high-frequency market noise characteristic of cryptocurrencies from their underlying long-term trends. Subsequently, the KAN-based mixers, with their ability to learn explicit non-linear functions, can model the complex dynamics within these separated components more effectively than traditional MLP or attention mechanisms, which may struggle with the extreme volatility and non-stationarity of crypto data. The results validate that DecoKAN offers an effective integration of signal processing and interpretable deep learning for addressing challenging real-world forecasting tasks.

\subsection{Computational Efficiency Analysis}

\begin{figure}[!t]
  \centering
  \includegraphics[width=0.5\textwidth]{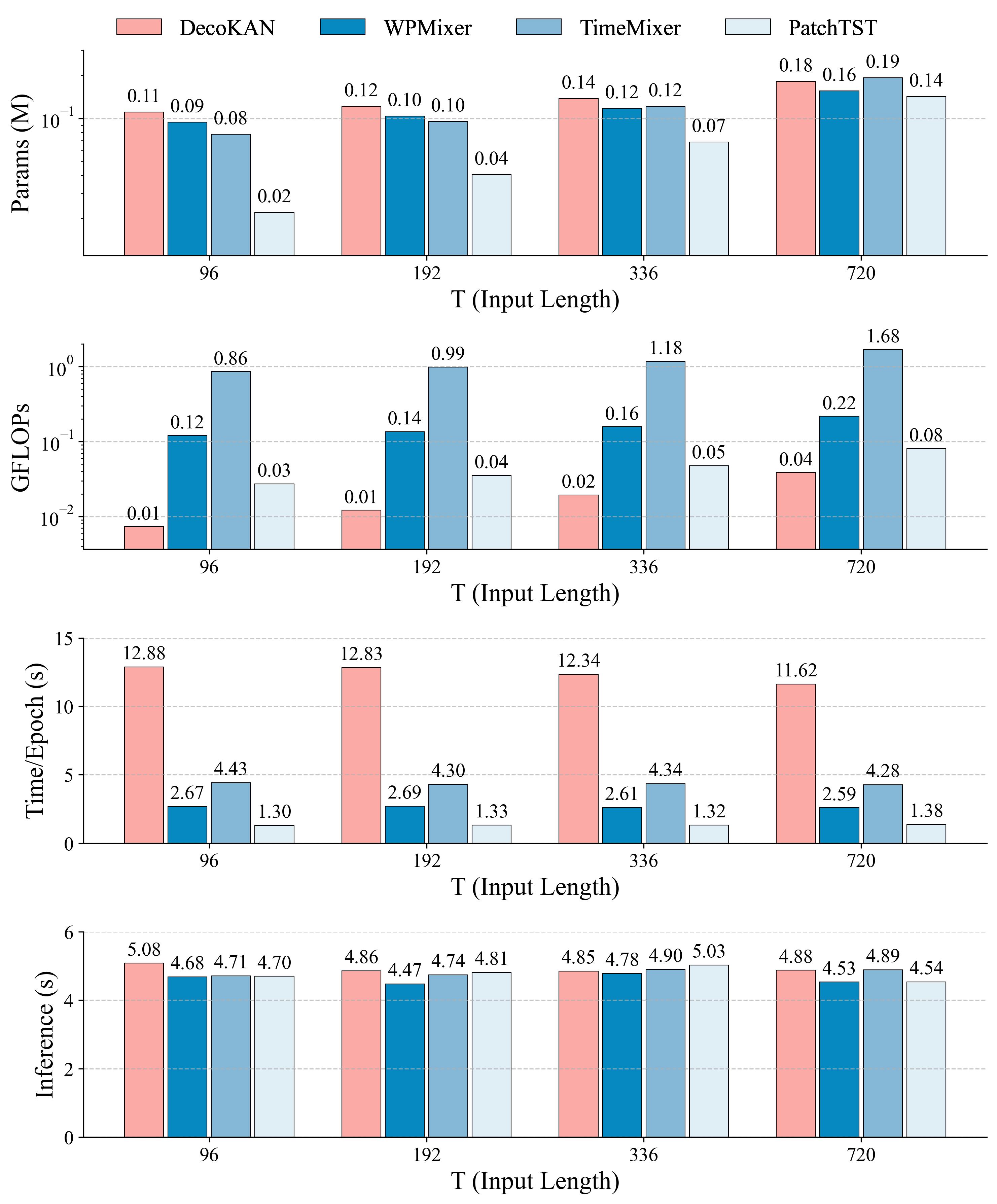}
  \caption{Computational efficiency comparison on the ETTh1 dataset ($L=96$). The plots illustrate (from top to bottom): total parameter count (M), theoretical computation in GFLOPs (log scale), empirical training time per epoch (s), and inference time (s) for various prediction lengths ($T$).}
  \label{fig:efficiency_comparison}
\end{figure}

The previous section examined DecoKAN's forecasting accuracy. To evaluate its practicality for real-world deployment, we analyze its efficiency from four complementary perspectives: parameter count, theoretical computation (GFLOPs), empirical training speed, and inference latency. All comparisons are conducted on the ETTh1 dataset ($L=96$) under unified configurations ($d_{model}=16$, Batch $=64$, $d_{ff}=4 \times d_{model}$) to ensure fairness. The results are summarized in Fig. 3.

\textbf{Parameter Count.} DecoKAN's parameter count ranges from 0.11M to 0.18M across different prediction horizons. This is broadly comparable to other MLP-based architectures such as WPMixer (0.09M–0.16M) and TimeMixer (0.08M–0.19M). The modest increase compared to the minimal PatchTST model follows naturally from the use of B-spline coefficients in the KANLinear layers, which introduce additional learnable parameters per connection to enable flexible non-linear modeling capabilities.

\textbf{Computation complexity (GFLOPs).} In terms of theoretical computational complexity, DecoKAN demonstrates exceptional efficiency. It achieves the lowest GFLOPs across all settings; for instance, at $T=96$, its 0.0073 GFLOPs is approximately 16 times lower than WPMixer (0.12) and orders of magnitude lower than TimeMixer (0.86). This highlights the intrinsic efficiency of the sparse KAN architecture in representing complex functions with fewer floating-point operations compared to dense MLP computations.

\textbf{Training Time.} Despite its low theoretical cost, DecoKAN exhibits the longest empirical training time. At $T=96$, each epoch takes approximately 12.6 seconds, which is about 4.9 times slower than WPMixer ($\approx 2.5$s) and nearly 10 times slower than PatchTST ($\approx 1.3$s). This discrepancy stems from the current implementation of B-spline computations, which are memory-intensive and less optimized for parallel GPU execution compared to the highly optimized dense matrix multiplications central to standard MLPs and Transformers.

\textbf{Inference Time.} Crucially for deployment, the inference latency of DecoKAN remains highly competitive. The inference time for the full test set (e.g., $\approx 5.1$s at $T=96$) is nearly identical to that of the baseline models, such as WPMixer (4.7s) and PatchTST (4.7s). 
This indicates that the computational overhead is primarily confined to the training phase and does not hinder the model's efficiency during real-time forecasting applications.
Nevertheless, the results confirm that the computational overhead is strictly confined to the offline training phase and does not hinder the model's efficiency during real-time forecasting applications.

\textbf{Summary.} The comprehensive efficiency analysis reveals that DecoKAN embodies a distinct computational trade-off. While it incurs higher training costs due to the current lack of hardware optimization for B-spline computations , this investment yields significant returns in forecasting accuracy (Sec. 4.2) and intrinsic symbolic interpretability (Sec. 5). 
Most importantly, DecoKAN's exceptionally low theoretical complexity translates into inference speeds that are fully viable for real-time deployment in high-frequency trading systems.

\subsection{Ablation Study}

To quantitatively validate the contribution of the core KAN components within the DecoKAN framework, we conducted a comprehensive ablation study on both a general benchmark (ETTh2) and a target-domain dataset (ETH). We tested the following four model configurations:

\begin{itemize}
\item \textbf{Full KAN:} The complete model, utilizing KANs for both temporal and feature mixing.
\item \textbf{Temporal KAN only:} A variant where the Feature KAN mixer is replaced by a standard MLP, isolating the contribution of the temporal mixing component.
\item \textbf{Feature KAN only:} A variant where the Temporal KAN mixer is replaced by an MLP, isolating the contribution of the feature mixing component.
\item \textbf{Baseline (MLP only):} A variant where KAN mixers are replaced by standard MLPs while preserving DecoKAN's exact backbone and hyperparameters. Note that this internal baseline is architecturally distinct from the WPMixer \cite{murad2025wpmixer} model, designed specifically here to strictly isolate the contribution of the KAN mechanism.
\end{itemize}

For each configuration, we conducted five independent runs using different random seeds to ensure the robustness of our findings. The distribution of the total average Mean Squared Error (MSE) for each case is presented in the box plots in Fig. 4.

\begin{figure}[!t]
  \centering
  \includegraphics[width=0.5\textwidth]{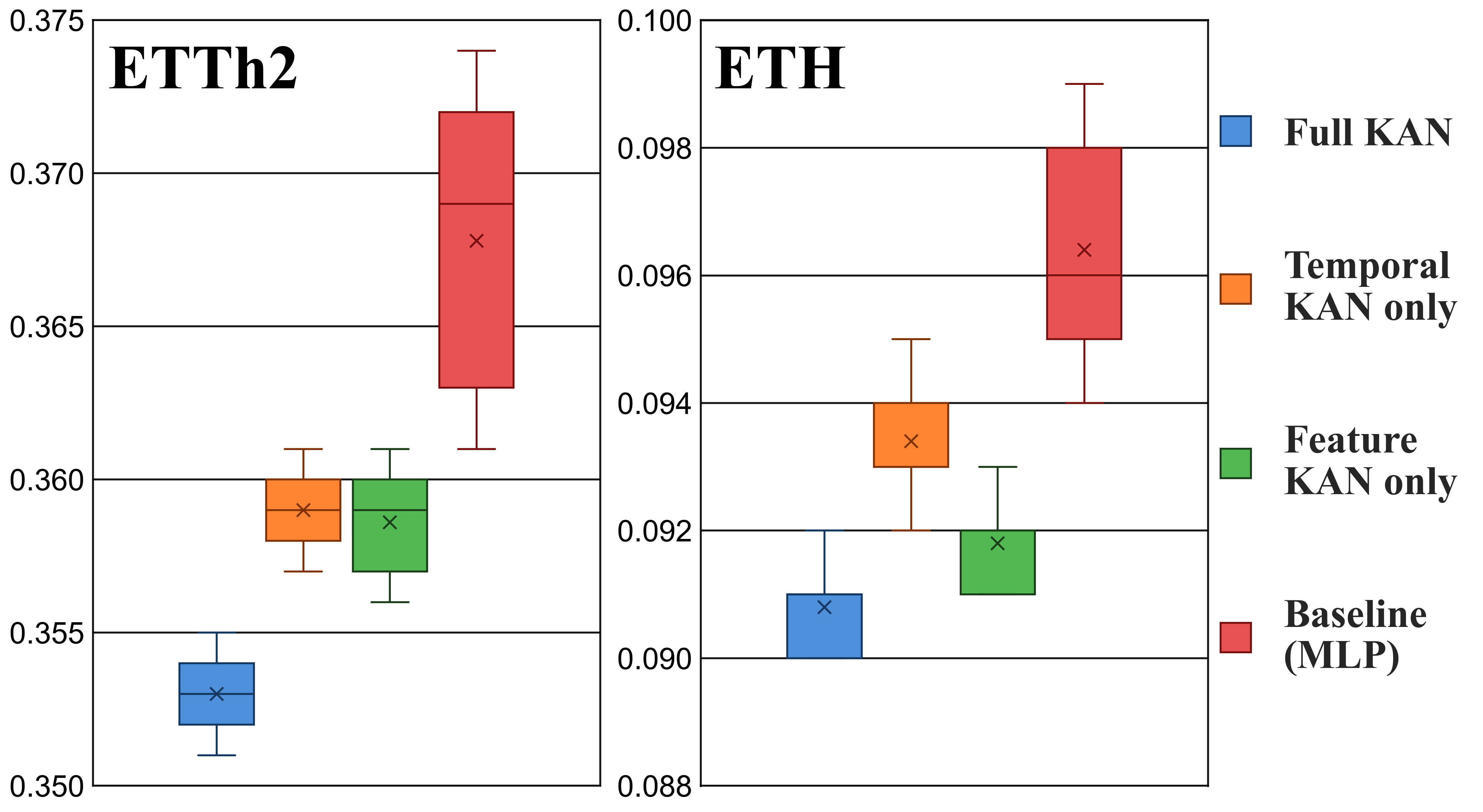}
  \caption{Ablation study results on the ETTh2 and ETH datasets. Each box plot shows the distribution of average MSE over 5 independent runs with different random seeds. The 'x' marks the mean performance. Lower values indicate better performance, and a smaller box indicates higher stability.}
  \label{fig:ablation_study}
\end{figure}

\textbf{Analysis of ETTh2 Dataset:}
On the ETTh2 dataset, the results (Fig. 4, left panel) show a general performance hierarchy. The Full DecoKAN model achieves the lowest average MSE. Removing either the temporal KAN (Temporal KAN only) or the feature KAN (Feature KAN only) component leads to a noticeable increase in average MSE, with both variants performing similarly to each other but worse than the full model. 
The Baseline (MLP only) model yields the highest average MSE, 
Since this baseline shares the identical decomposition structure as DecoKAN, this performance gap directly validates the superior modeling capability of KAN mixers over standard MLPs, independent of the wavelet framework itself.
This validates that for a general, seasonal time series like ETTh2, both the temporal and feature KAN components contribute positively to achieving optimal performance.

\textbf{Analysis of ETH Dataset:}
Notably, the ablation study on the volatile ETH dataset reveals a different dynamic (Fig. 4, right panel). The performance distribution of Feature KAN only closely matches that of the Full DecoKAN model, indicating similar average MSE and variance across runs. In contrast, the Temporal KAN only configuration exhibits a clear increase in average MSE compared to the Full DecoKAN. The Baseline (MLP only) configuration yields the highest average MSE among the tested variants on this dataset as well.

This distinct behavior aligns with the inherent characteristics of cryptocurrency markets. Unlike traditional time series (e.g., electricity load) dominated by strong temporal autocorrelation, cryptocurrency prices are often driven by high-frequency shocks and the instantaneous interplay between system variables (e.g., the non-linear coupling between trading volume, transaction fees, and price volatility). Consequently, the performance of DecoKAN here relies predominantly on the Feature KAN component's ability to capture these complex cross-variate dependencies, proving more critical than the temporal mixing for this domain. This indicates DecoKAN can adapt the relative importance of its internal components based on data characteristics. Furthermore, the observed performance difference between the KAN-based configurations and the MLP baseline, coupled with DecoKAN's intrinsic interpretability (as demonstrated in Section 4.5), underscores its advantages for trustworthy forecasting in critical financial applications.

The compact nature of the box plots across both datasets also visually confirms the stability and robustness of the DecoKAN architecture against random initializations.

\subsection{Interpretability Analysis}

A core advantage of DecoKAN is its intrinsic interpretability. This is realized by leveraging the three-stage pipeline native to Kolmogorov-Arnold Networks conceptually illustrated in Fig. 5—Sparsification, Pruning, and Symbolification—which allows transforming the trained network's learnable activation functions into concise, mathematically explicit symbolic formulas. Sparsification is achieved during training via the regularization loss $L_{\rm reg}$ (Eq. 12), encouraging parsimonious solutions. This section presents a case study on the ETH price prediction task to demonstrate how subsequent Pruning and Symbolification reveal the internal functional specialization of the DecoKAN framework.

\begin{figure}[!t]
\centering
\includegraphics[width=0.5\textwidth]{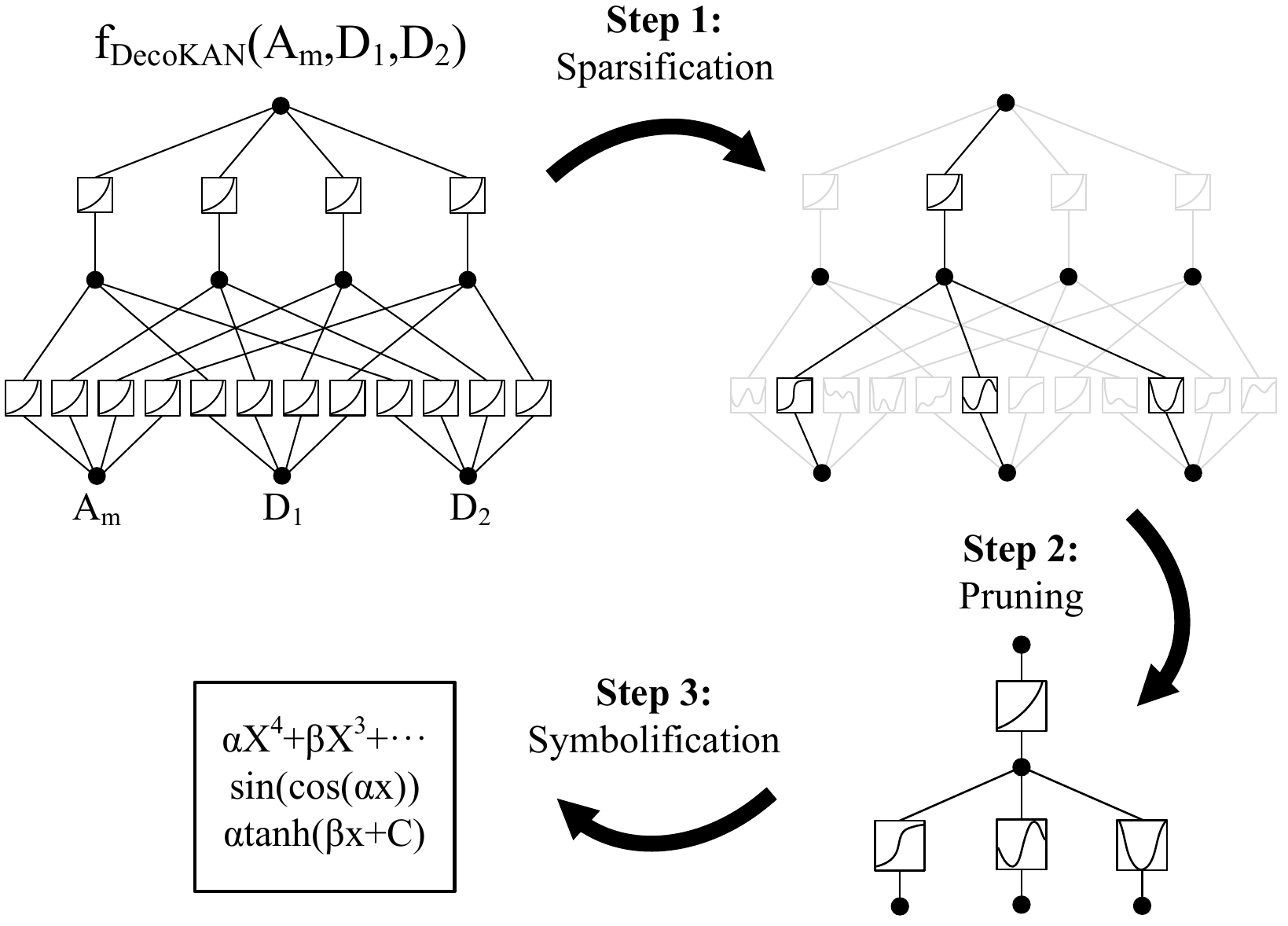}
\caption{An example of how to do symbolic regression with KAN. Steps 1--3 correspond to Lines 21, 34, 35 of Algorithm 1 respectively.}
\label{fig:kan_symbolic}
\end{figure}

\begin{table}[H]
\renewcommand{\arraystretch}{1.2}
\setcounter{table}{3}
\caption{Pruning statistics of DecoKAN on the ETH dataset ($\tau = 0.05$). Values denote the count of learnable edges.}
\label{tab:pruning_status}
\centering
\footnotesize
\begin{tabular}{l|c|c|c|c}
\toprule
\textbf{Branch} & \textbf{Total} & \textbf{Pruned} & \textbf{Preserved} & \textbf{Prune Ratio} \\
\midrule
\textbf{Approximation} & 12,720 & 609 & 12,111 & 4.79\% \\
\textbf{Detail} & 12,720 & 9,703 & 3,017 & 76.28\% \\
\textbf{Overall Model} & 25,440 & 10,312 & 15,128 & 40.53\% \\
\bottomrule
\end{tabular}
\end{table}

\begin{table*}[!b]
\renewcommand{\arraystretch}{1.2}
\caption{Top Symbolic Formulas Extracted from KAN Mixer Connections on ETH Dataset (Ranked by $R^2$ Score)
(Layer i and Layer j denote the input and output node indices of the respective KAN Linear layer within the specified Mixer block)}
\label{tab:symbolic_results}
\centering
\footnotesize
\setlength{\tabcolsep}{3.317mm}
\begin{tabular}{l|c|c|c|c|c}
\toprule
\textbf{Branch} & \textbf{KAN Mixer} & \textbf{Layer \textit{i}} & \textbf{Layer \textit{j}} & \textbf{Symbolic Formula} & \textbf{R² Score} \\
\midrule
Detail Branch  & KAN Mixer 2  & 120 & 15 & $-0.076x^4 + 0.216x^3 + 0.252x^2 - 1.370x - 0.116$ & 0.9968 \\
Detail Branch  & KAN Mixer 1  & 126 & 8 & $-0.153x^4 + 0.170x^3 + 0.610x^2 - 1.270x - 0.334$ & 0.9958 \\
Detail Branch  & KAN Mixer 1  & 111 & 2 & $-1.403\sin(0.905x + 0.039) - 0.156\cos(2.721x)$ & 0.9941 \\
Detail Branch  & KAN Mixer 1  & 43 & 26 & $-1.307\sin(1.072x + 0.056) - 0.186\cos(-3.039x)$ & 0.9937 \\
Detail Branch  & KAN Mixer 1  & 20 & 72 & $-0.047x^4 + 0.226x^3 + 0.080x^2 - 1.386x + 0.004$ & 0.9931 \\
Detail Branch  & KAN Mixer 2  & 36 & 11 & $0.163x^4 - 0.195x^3 - 0.596x^2 + 1.317x + 0.302$ & 0.9926 \\
Detail Branch  & KAN Mixer 1  & 8 & 4 & $-1.282\tanh(1.431x + 0.029)$ & 0.9905 \\
Detail Branch  & KAN Mixer 1  & 72 & 2 & $0.112x^4 - 0.144x^3 - 0.390x^2 + 1.240x + 0.189$ & 0.9884 \\
Approx Branch  & KAN Mixer 2  & 26 & 4 & $0.923\sin(1.348x + 0.695) + 0.801\cos(2.624x)$ & 0.9862 \\
Detail Branch  & KAN Mixer 1  & 23 & 1 & $1.044\sin(1.312x + 0.462) + 0.678\cos(2.631x)$ & 0.9826 \\
\bottomrule
\end{tabular}
\end{table*}

For this case study, we employ a single-level wavelet decomposition ($m=1$). This configuration simplifies the model into two distinct processing pathways, each with a dedicated KAN Resolution Branch:

\begin{itemize}
\item \textbf{Approximation Branch}, processing the low-frequency approximation coefficients ($A_1$) related to the underlying trend.
\item \textbf{Detail Branch}, processing the high-frequency detail coefficients ($D_1$) related to volatility and noise components.
\end{itemize}

Following sparsification-guided training, the network is pruned by removing connections whose spline activation function's L2 norm falls below a threshold $\tau$ (set to 0.05 in this study). Table IV summarizes the pruning status for the two branches, revealing their structural efficiency.

The pruning results in Table IV highlight a clear functional specialization. The Approximation Branch exhibits a low pruning ratio of only 4.79\%, signifying that its connections are fundamental and highly efficient, forming an indispensable structural backbone for preserving the low-frequency components of the ETH prices. Conversely, the Detail Branch shows a much higher pruning ratio of 76.28\%, indicating a high degree of learned sparsity. 
This suggests that while the branch begins with high capacity, it effectively selects only the most critical connections to model the transient fluctuations.

Finally, symbolification replaces the remaining B-spline activation functions in the pruned network with mathematically explicit symbolic formulas, selected based on the best fit ($R^2$) from a candidate library (e.g., polynomials, trigonometric). To demonstrate the model's capability to discover explicit physical dynamics from data, Table V lists the top symbolic functional relationships discovered by the DecoKAN model on the ETH dataset, rigorously ranked by their coefficient of determination ($R^2$).

This detailed symbolic analysis corroborates the functional specialization observed during pruning, revealing three critical insights within a unified view. First, the consistently high $R^2$ scores (predominantly $>0.99$) across the top rankings validate the KAN mixers' convergence to precise functional mappings. Second, the Detail Branch overwhelmingly dominates these rankings with complex polynomial and trigonometric formulas (e.g., Rank 3), providing explicit mathematical explanations for high-frequency market volatility. In contrast, the Approximation Branch appears sparsely in this high-confidence symbolic list (represented solely by Rank 9). This scarcity aligns with our pruning analysis, suggesting that the Approximation Branch functions primarily as a stable, dense structural backbone that preserves low-frequency components through distributed weights rather than relying on specific symbolic curve-fitting.

In summary, the interpretability analysis validates DecoKAN's design, revealing a clear functional division: the Approximation Branch provides a robust structural backbone for the forecasting task, while the Detail Branch acts as a high-capacity engine for capturing complex volatility dynamics.

\section{Conclusion}

This paper presents DecoKAN, a framework combining multi-resolution wavelet decomposition (DWT) with the intrinsic interpretability of Kolmogorov–Arnold Networks (KANs) to generate explicit mathematical laws from data for multivariate time series forecasting in cryptocurrency systems. Through extensive experiments, DecoKAN demonstrated competitive predictive accuracy and improvements in forecasting volatile cryptocurrency markets. Interpretability analysis revealed a functional separation within the model: the Detail Branch captures explicit symbolic relations for high-frequency fluctuations, 
while the Approximation Branch acts as a robust structural backbone for the forecasting task.

However, limitations remain. A primary constraint is computational efficiency. KANs rely on B-spline computations that are currently less optimized for GPU parallelization, leading to slower training times compared to MLP or Transformer-based architectures. 
Crucially, however, this overhead is confined to training; our experiments confirm that DecoKAN's inference latency remains highly competitive, ensuring its practical viability for real-time deployment.
Additionally, the symbolification process depends on a finite set of candidate functions, and interpreting the resulting symbolic graphs requires human domain expertise.

Future work will address these challenges. 
Methodologically, we aim to bridge the training efficiency gap by incorporating recent advances in efficient KAN implementations, such as CUDA-accelerated grids and network quantization, to significantly reduce overhead without compromising interpretability.
Architecturally, integrating KAN modules into other forecasting backbones may enable new interpretable models for time series. 
In applications, DecoKAN's transparency makes it suitable for analyzing collective market behavior and social sentiment dynamics, including market manipulation detection, algorithmic trading, and risk assessment in DeFi.

\bibliographystyle{IEEEtran}
\bibliography{references}

\begin{IEEEbiography}[{\includegraphics[width=1in,height=1.25in,clip,keepaspectratio]{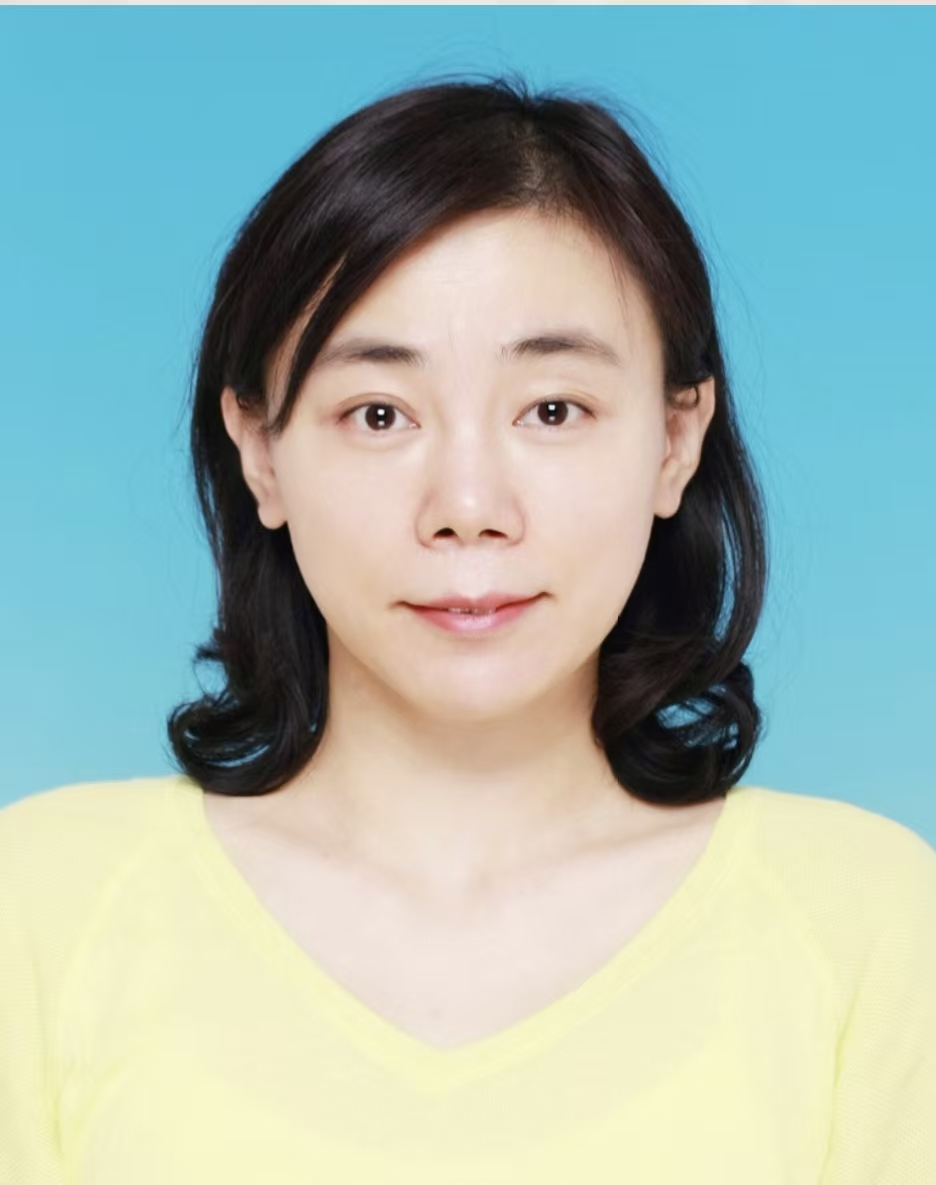}}]{Yuan Gao}
received the Ph.D. degree in computer application technology from the Beijing University of Technology in 2016. She completed her postdoctoral research at the State Information Center in 2024. She is currently an Associate Professor with the Department of Electronics and Communication Engineering, Beijing Electronic Science and Technology Institute. Her research interests include blockchain, big data analysis, visualization and computer graphics, and model security.
\end{IEEEbiography}

\begin{IEEEbiography}[{\includegraphics[width=1in,height=1.25in,clip,keepaspectratio]{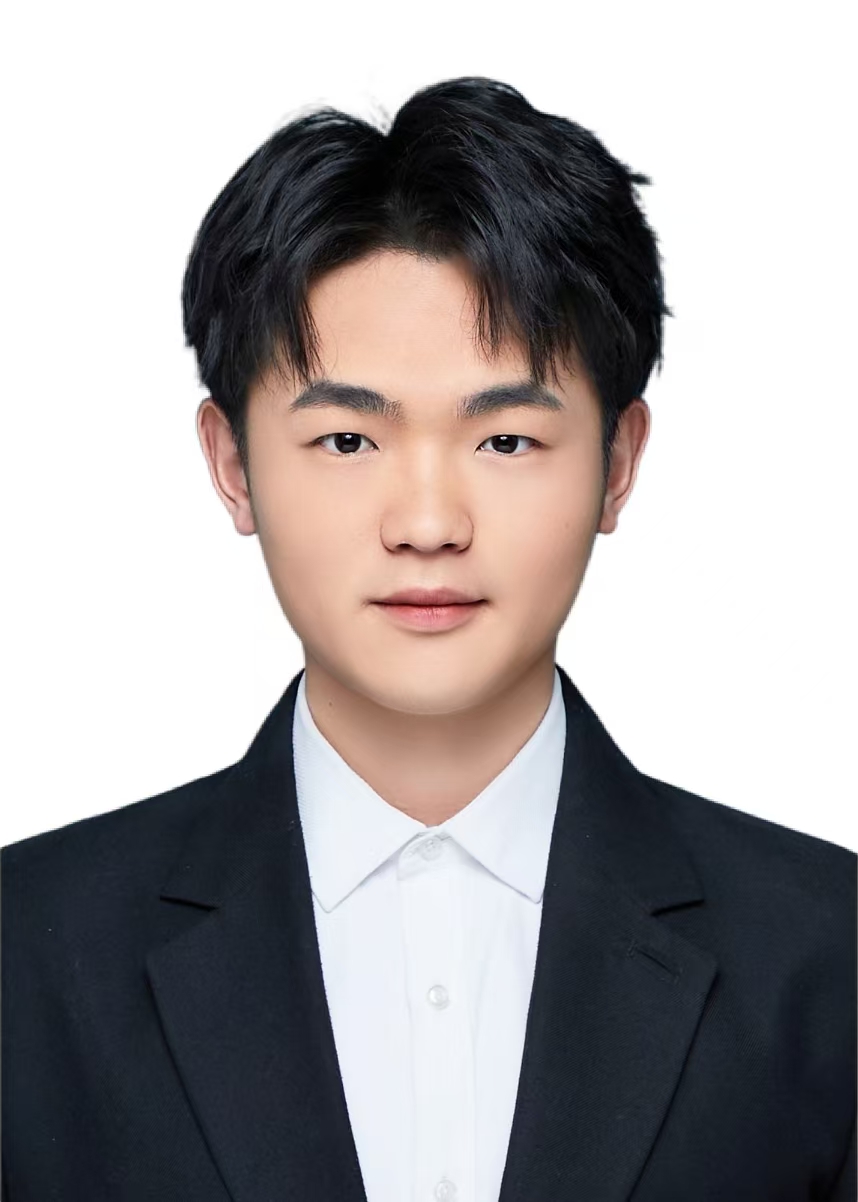}}]{Zhenguo Dong}
received the B.S. degree in Computer Science and Technology from the Shandong University of Technology in 2024. He is currently a graduate student with the Beijing Electronic Science and Technology Institute. His research interests include blockchain, multivariate time series forecasting, and model security.
\end{IEEEbiography}

\begin{IEEEbiography}[{\includegraphics[width=1in,height=1.25in,clip,keepaspectratio]{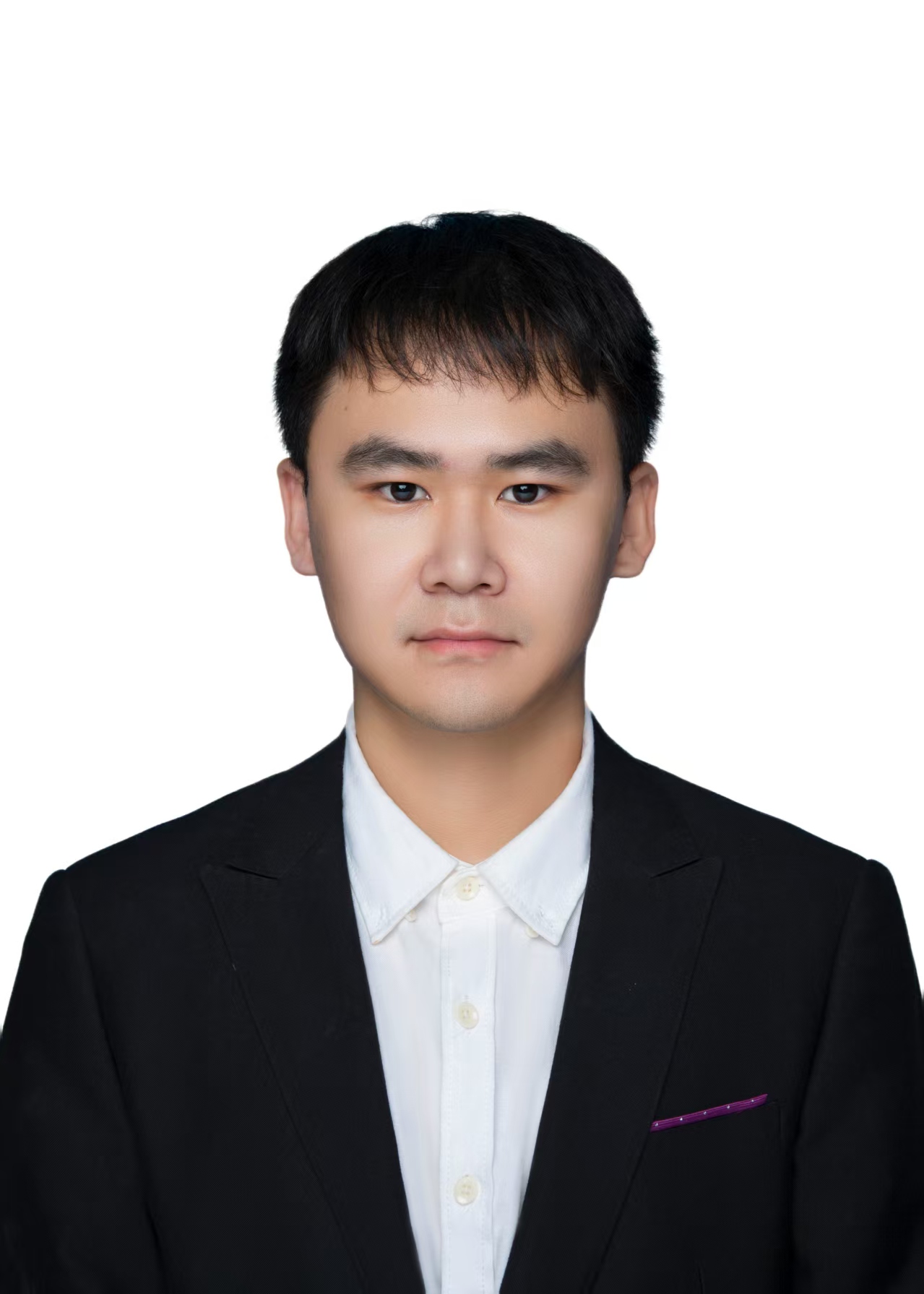}}]{Xuelong Wang}
received the B.S. degree in microelectronic science and engineering from the Hefei University of Technology in 2023. He is currently a graduate student with the Beijing Electronic Science and Technology Institute. His research interests include time series data analysis and model security.
\end{IEEEbiography}

\begin{IEEEbiography}[{\includegraphics[width=1in,height=1.25in,clip,keepaspectratio]{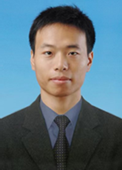}}]{Zhiqiang Wang}
received the Ph.D. degree in Information Security from the School of Communication Engineering, Xidian University in 2015. He is currently an Associate Professor with the Department of Cyberspace Security, Beijing Electronic Science and Technology Institute. His research interests include privacy protection and cyberspace security.
\end{IEEEbiography}

\begin{IEEEbiography}[{\includegraphics[width=1in,height=1.25in,clip,keepaspectratio]{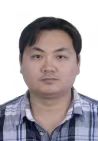}}]{Yong Zhang}
(Member, IEEE) received the Ph.D.degree in computer science from the Beijing University of Technology in 2010. He is currently a Professor in computer science with the Beijing University of Technology. His research interests include intelligent transportation systems, big data analysis, visualization, and computer graphics.
\end{IEEEbiography}

\begin{IEEEbiography}[{\includegraphics[width=1in,height=1.25in,clip,keepaspectratio]{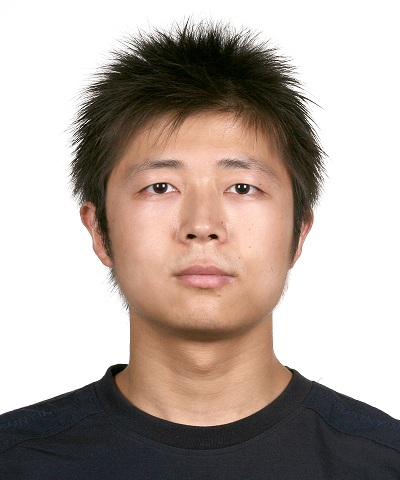}}]{Shaofan Wang}
received the B.S. and Ph.D. degrees in  mathematics from  Dalian University of Technology, Dalian, China in 2003 and 2010, respectively. He is an associate professor from the Faculty of Information Technology, Beijing University of Technology, Beijing, China. His research interest includes pattern recognition and computer vision.
\end{IEEEbiography}

\end{document}